\documentclass[10pt, a4paper]{article}

\usepackage{lrec-coling2024} % this is the new style
\usepackage{multirow}
\usepackage{makecell}

\title{\textsc{REFeREE}: A REference-FREE Model-Based Metric \\ for Text Simplification}

\name{Yichen Huang, Ekaterina Kochmar} 

\address{Mohamed bin Zayed University of Artificial Intelligence \\
         \{yichen.huang, ekaterina.kochmar\}@mbzuai.ac.ae\\}

\abstract{
Text simplification lacks a universal standard of quality, and annotated reference simplifications are scarce and costly. We propose to alleviate such limitations by introducing \textsc{REFeREE}, a \textsc{re}ference-\textsc{free} model-based metric with a 3-stage curriculum. \textsc{REFeREE} leverages an arbitrarily scalable pretraining stage and can be applied to any quality standard as long as a small number of human annotations are available. Our experiments show that our metric outperforms existing reference-based metrics in predicting overall ratings and reaches competitive and consistent performance in predicting specific ratings while requiring no reference simplifications at inference time.
 \\ \newline  }

\begin{document}

\maketitleabstract

\section{Introduction}

The task of text simplification (TS) aims to reduce the reading and grammatical complexity of text while retaining its meaning and grammaticality~\cite{chandrasekar1997automatic}. Alongside other text-to-text generation tasks (such as machine translation or summarization), it has been a common practice in TS to use reference-based automatic metrics and evaluate a generated text by comparing it to gold-standard references, typically produced by humans. However, the complexities of TS pose particular challenges for this type of evaluation approaches, and prevalent metrics such as BLEU~\cite{bleu}, SARI~\citep{sari} and BERTScore~\cite{bertscore} have been shown to correlate poorly with human evaluation~\cite{simplicity_da_eval, lens}. 

To a large extent, this is due to the absence of a singular, precise definition of what text simplification aims to do and how the output quality should be judged: text simplification may involve lexical, syntactic and conceptual modifications conducted using different operations (e.g. word-swapping, sentence-splitting, and paraphrasing)~\cite{sulem-etal-2018-semantic}, and simplification quality is associated with different aspects such as fluency (or grammaticality), meaning preservation (or adequacy) and simplicity. Since reference-based metrics rely on availability of a large enough number of diverse yet high-quality references, the availability of such reference outputs creates a clear bottleneck, as collection of human-produced references is a slow and expensive process. Even though some high quality TS corpora (e.g., Newsela~\citeplanguageresource{newsela}) exist, they are still costly to create and often are not open-access~\cite{martin-etal-2018-reference}. Finally, there are cases when such references will be impossible to collect: e.g., when there is a need to estimate the quality of text in real time as is increasingly becoming the case for texts generated using large language models (LLMs)~\cite{zhang-et-al}. 

The above reasons motivate development of supervised, model-based evaluation approaches, where a model is trained to mimic human evaluation on given examples, which is applicable to any standard as long as the annotations are available and internally consistent. The need for reliable reference-free evaluation metrics has been expressed before~\cite{specia2010machine,thompson-post-2020-automatic}, and more recently a number of learnable TS metrics have been proposed~\cite{lens,bets,sle}. For instance,~\citet{lens} have proposed LENS, where they use RoBERTa-extracted~\cite{roberta} representations of (source, simplification, reference) tuples to predict overall quality scores as annotated by humans. Whereas LENS significantly outperforms conventional metrics in correlation with human judgements, it is severely limited by the scarcity of human annotations, with its training data consisting of only 2.4K simplification outputs from 24 systems. 

Inspired by the success of BLEURT~\cite{bleurt} in machine translation, we propose pretraining on synthesised data and supervision signals as a means to leverage large-scale, unlabeled data and overcome the bottleneck of reference simplifications and human ratings. To facilitate the arbitrarily scalable synthesis of pretraining data, we argue for a reference-free, source-based metric. Specifically, we use existing TS models to produce simplifications for arbitrary source sentences. Given (source, simplification) pairs, we task a model-based metric to predict a range of synthesised supervision signals such as BERTScore, GPT-2 perplexity and model-based simplicity ratings. An additional benefit is that, by enabling direct comparison with the source sentence, such a metric can more accurately evaluate criteria such as meaning preservation and relative simplicity (as opposed to source-free metrics such as BLEURT and BERTScore).

\begin{figure}[t]
\begin{center}
\includegraphics[width=0.48\textwidth]{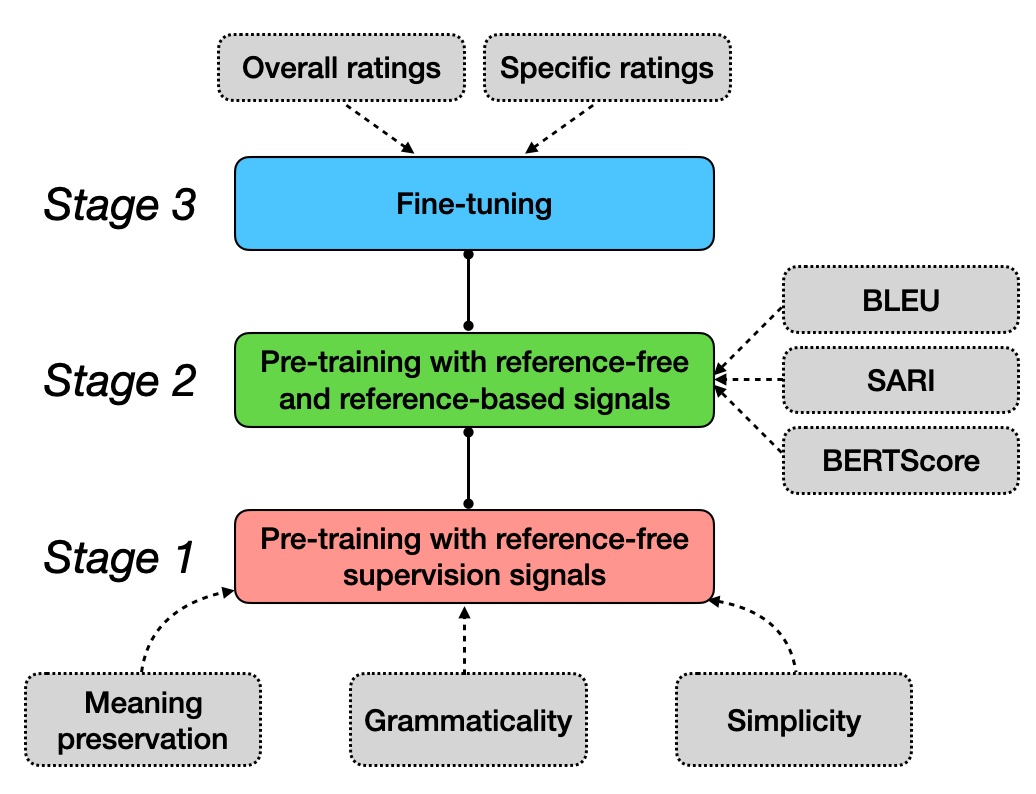} 
\end{center}
\caption{An overview of the proposed curriculum.}
\label{fig:overview}
\end{figure}

In this work, we introduce \textsc{REFeREE}, a model-based metric for text simplification that is reference-free. We propose a curriculum with two pretraining stages and a fine-tuning stage as shown in Figure \ref{fig:overview}. The first pretraining stage uses reference-free supervision signals and is arbitrarily scalable, allowing us to leverage the large amounts of unlabeled texts. The second pretraining stage relies on both reference-free and reference-based supervision signals. This stage makes use of the readily available TS corpora that do not include human ratings and provides more accurate supervision. Finally, we fine-tune the metric on human ratings such that it is aligned with the specific simplification operations and criteria.

We evaluate our approach on overall ratings from the \textsc{SimpEval} dataset \citeplanguageresource{lens} and specific adequacy, fluency and simplicity ratings from the smaller \textsc{Simplicity-DA} \citeplanguageresource{simplicity_da} and \textsc{Human-Likert} \citeplanguageresource{scialom2021rethinking} datasets. \textsc{REFeREE} correlates better with the overall human ratings, outperforming popular rule-based metrics, BERTScore, BLEURT, LENS and other model-based strong baselines while using a smaller model and requiring less information at inference time. On the smaller \textsc{Simplicity-DA} and \textsc{Human-Likert} datasets, \textsc{REFeREE} overall underperforms \textsc{Lens} but still performs better than conventional metrics and is more consistent across different datasets. Additionally, we perform extensive ablation studies to investigate the effects of each component in our training process.\footnote{Code and model checkpoints are available at \url{https://github.com/i-need-sleep/referee}.}

\section{Related Work}

The literature on TS models and automatic evaluation is vast. In this section, we provide a brief outline of the types of model-based metrics and then give a more detailed account of the previous work that addresses reference-free evaluation.

\subsection{Model-Based Evaluation Metrics}
Model-based evaluation metrics have been actively investigated in NLP and particularly machine translation (MT). Metrics fine-tuned on human ratings, such as BLEURT \cite{bleurt, bleurt_scaled_up}, COMET \cite{comet}, and \textsc{UniTE} \cite{unite}, have been shown to produce good results. In this work, we take direct inspiration from BLEURT, a generic metric for natural language generation (NLG). BLEURT produces evaluation from (prediction, reference) pairs and utilizes an arbitrarily scalable pretraining stage where it is trained to predict automatically generated supervision signals (BLEU, BERTScore, etc.) based on synthesized, semantically similar pairs. The pretraining step allows BLEURT to produce good results even when fine-tuned on a reasonably small set of human annotations. However, this method is not immediately applicable to TS evaluation as reference simplifications are difficult to synthesize and TS evaluation involves particular aspects of quality (e.g. simplicity) not considered in BLEURT's pretraining objectives. In response to these challenges, we propose a reference-free, source-based setup with a modified pretraining process.

There also exists a family of unsupervised metrics that do not rely on human ratings and instead compare the embeddings either between the system output and the reference \cite{bertscore} or between the system output and the source sentence \cite{xmoverscore, distilscore, uscore}. This line of research is not immediately suitable for our task as text simplification can be evaluated in terms of different aspects, resulting in different ratings even when given the same (source, simplification, reference) tuple.

Another recent trend that addresses automatic evaluation is based on the direct use of Large Language Models (LLMs) for evaluation. \citet{gptscore} use LLM-predicted conditional probabilities as estimates for NLG quality. \citet{g_eval} task LLMs to directly produce numeric ratings given descriptions of evaluation criteria through a chain-of-thought \cite{cot} process. With natural language instruction as an interface for defining evaluation criteria, this type of approach is flexible, data-lean at inference time, and has been shown to correlate well with human ratings. However, LLMs are compute-hungry and tend to overestimate outputs generated by models similar to themselves \cite{g_eval}. In addition, most existing LLM-based metrics are aligned with human raters implicitly (through natural language instruction) rather than explicitly (with human ratings). As such, we see this line of work as orthogonal to ours.

\subsection{Reference-Free Evaluation Metrics}
Traditionally, the outputs of text-to-text generation models have been evaluated using {\em reference-based metrics} such as BLEU~\cite{bleu}, ROUGE~\cite{rouge}, or BERTScore~\cite{bertscore}. These metrics do not always correlate with human judgments of the generated output quality~\cite{sulem-etal-2018-bleu}, and such evaluation paradigm falls short when human references are not available or only a single model is available~\cite{louis2013automatically}. The quality of the references also has an impact on reference-based evaluation, and, given that  evaluation of generated outputs is cognitively demanding and highly subjective~\cite{vasilyev-etal-2020-fill}, it is hard to avoid variability across a set of references and gold standard judgements~\cite{harman-over-2004-effects}.

As a result, development of Quality Estimation (QE)~\cite{specia2010machine} and {\em reference-free metrics} has gained increased attention in the recent years~\cite[inter alia]{louis2013automatically,scialom-etal-2019-answers,thompson-post-2020-automatic}.
For summarization, ~\citet{louis2013automatically} show that quantifying the similarity between the source text and its summary with appropriately chosen content similarity based measures produces scores which replicate human assessments accurately. \citet{scialom-etal-2019-answers,scialom-etal-2021-questeval} and~\citet{vasilyev-etal-2020-fill} evaluate summary quality by measuring how such summaries help with related tasks rather than how well they align with a pre-defined set of references: 
\citet{scialom-etal-2019-answers,scialom-etal-2021-questeval} introduce metrics based on the intuition that the quality of a generated summary is directly related to the number of relevant questions that can be answered on its basis; and~\citet{vasilyev-etal-2020-fill} estimate summary quality measuring the performance boost gained by a pre-trained language model with access to the summary while carrying out its language understanding task on the document's text. 
In MT, such systems as COMET and its extensions, which use pre-training and subsequent normalization, and Prism~\cite{thompson-post-2020-automatic}, which casts the evaluation task to that of scoring MT output with a zero-shot paraphraser, show results competitive with reference-based models~\cite{comet,rei-etal-2021-references}; finally,~\citet{fonseca-etal-2019-findings} demonstrate that such metrics also highly correlate with human judgments while alleviating the need for references, thus suggesting that reference-free evaluation is a promising direction for future research.

\subsection{Reference-Free TS Evaluation}

In TS, researchers have also been investigating application of reference-free measures~\cite[inter alia]{c-score,kajiwara-fujita-2017-semantic}: for instance, \citet{c-score} propose to evaluate TS quality extrinsically via human reading comprehension. Other early works on automatic reference-free evaluation rely on feature-based classification approaches aimed at specific aspects of simplification (e.g., lexical, syntactic, structural). For instance, \citet{stajner-etal-2014-one} investigate applicability of popular MT evaluation metrics to the TS systems outputs and the corresponding original sentences, and demonstrate their potential in replacing human assessment of TS systems aimed at syntactic simplifications and content reduction. \citet{kajiwara-fujita-2017-semantic} show that a classification model utilizing alignment-based semantic features is capable of reliably predicting TS system quality when lexical simplifications are also involved as is the case with the QATS 2016 dataset~\cite{vstajner2016shared}. Experiments of~\citet{martin-etal-2018-reference} on the same data demonstrate that n-gram-based MT metrics correlate the most with human judgment of grammaticality and meaning preservation, whereas simplicity is best evaluated by basic length-based metrics. Finally,~\citet{sulem-etal-2018-semantic} focus on the structural aspects of TS and propose SAMSA, a system that uses decomposition of the input based on its semantic structure and compares it to the output.

A different line of research investigates the use of large-scale pre-trained models for direct quality estimation in TS systems: a notable example is~\citet{kriz2020simple}, who propose Simple-QE, a BERT-based QE model adapted from prior summarization work, and show that it correlates well with human quality judgments. \citet{scialom2021rethinking} present a BERTScore-based~\cite{bertscore} adaptation of their {\sc QuestEval} metric~\cite{scialom-etal-2021-questeval} for TS and show that it yields competitive results along the meaning preservation dimension, with considerable improvement over BLEU and SARI. At the same time, their analysis suggests that datasets commonly considered in TS research show a considerable level of spurious correlations between different dimensions, with fluency being highly correlated with meaning preservation and simplicity. To that end, they release {\sc Human-Likert}, a new large corpus of human evaluations devoid of such spurious correlations. 

\begin{table*}[ht]
\centering
\resizebox{2\columnwidth}{!}{%
\begin{tabular}{lll}
\hline
\textbf{Training stage} & \textbf{Data} & \textbf{Supervision signals} \\
\hline
Pretraining (Stage 1) & OpenWebText & \makecell[l]{ \textbf{Meaning preservation}: SBERT, self-BLEU, self-BERTScore \\ \textbf{Fluency}: Source perplexity, simplification perplexity \\ \textbf{Simplicity}: Source FKGL, simplification FKGL, \\ \hspace{15 mm} source simplicity, simplification simplicity} \\
\hline
Pretraining (Stage 2) & \makecell[l]{Newsela (test) \\ WikiSmall (test) \\ WikiLarge (test)} & \makecell[l]{All stage 1 objectives \\ + BLEU, SARI, BERTScore} \\
\hline
Fine-tuning (Stage 3) & \makecell[l]{\textbf{Overall}: \textsc{SimpEval} \\ \textbf{Specific}: \textsc{Simplicity-DA}, \textsc{Human-Likert} } & Human ratings \\
\hline
\end{tabular}
}
\caption{An overview of the data and supervision signals used in each training stage. For the final stage, we fine-tune and evaluate \textsc{REFeREE} separately for each dataset and quality aspect.}
\label{tab:overview}
\end{table*}

Recently, \citet{bets} have proposed BETS, a reference-free TS metric aggregating a simplicity score and an adequacy score. The simplicity branch is trained on pairs of complex and simple phrases, and the adequacy branch is based on word embedding similarity akin to BERTScore. At the same time, \citet{sle} propose to evaluate TS quality with SLE, a reference-free simplicity metric trained on softened reading levels from Newsela \citeplanguageresource{newsela}. These metrics focus on specific simplification qualities and evaluation criteria: BETS primarily considers lexical simplification, with its simplicity branch unable to return a score when all words in the simplified sentence are present in the original sentence (as would be the case with simplification by deletion or splitting), and SLE is not trained nor tested on machine simplifications with adequacy or fluency issues. 

By contrast, we propose a metric that is compatible with any simplification operation as it is fine-tuned end-to-end on human ratings. To the best of our knowledge, the only learnable model-based TS metric aligned with human ratings is LENS \cite{lens}, which adopts a COMET-like approach and relies on references.

\section{Methodology}

\textsc{REFeREE} is based on a pretrained {\tt DeBERTa-v3-base} model \cite{debertav3} and takes as input delimited pairs of source sentences and system outputs. The DeBERTa-extracted sequence embedding is passed into a linear regression head for each supervision signal. The training process consists of three stages: (1) an arbitrarily scalable pretraining stage with reference-free supervision signals, (2) a second pretraining stage with both reference-free and reference-based supervision signals, and (3) a fine-tuning stage with human ratings. An overview of the data and supervision signals is shown in Table \ref{tab:overview}. We describe the three stages in the following subsections. More implementation details are reported in Appendix \ref{sec:implementation_details}.

\subsection{Pretraining with Reference-Free Supervision Signals}

The first pretraining stage aims to learn important aspects of simplification quality directly from text. It involves a collection of reference-free supervision signals and is designed to be arbitrarily scalable such that the metric can leverage large amounts of unlabeled text. Based on the common criteria of text simplification quality~\cite[inter alia]{martin-etal-2018-reference,kriz2020simple,scialom2021rethinking}, we select a range of supervision signals measuring meaning preservation (also referred as adequacy), fluency (or grammaticality) and simplicity:

\begin{itemize}
    \item {\bf Meaning preservation (adequacy)} focuses on {\em how well the TS output preserves the meaning of the original text}. For this, we use the cosine embedding distance from SBERT embeddings~\cite{sbert} as well as self-BLEU and self-BERTScore measured against the source sentence. In this way, our metric can capture both lexical overlaps and paraphrases. 
    \item {\bf Fluency (grammaticality)} aims to measure {\em how well-formed the TS output is}. For this, we include the perplexity for both the source sentence and the system output as measured by GPT-2~\cite{gpt2}. Our intuition is that if the content of the source and the system output is similar, then the difference between their perplexities reflects the difference in fluency. 
    \item {\bf Simplicity} focuses on {\em the extent to which the output is easier to read and understand than the original text}. Here, we utilize the FKGL score~\cite{fkgl} and a model-based readability score for both the source and the system output. We use an ALBERT-based~\cite{albert} system trained on the CommonLit dataset.\footnote{\url{https://github.com/mathislucka/kaggle_clrp_1st_place_solution}}
\end{itemize}

To generate pairs of source sentences and machine simplifications, we use a small subset of the OpenWebText dataset \citeplanguageresource{openwebtext} of approximately 200K sentences. To obtain a range of good and bad simplifications, we use outputs from high-performing models and augment the dataset with degraded simplifications. We use outputs from MUSS \cite{muss} as well as 5-shot results from GPT-3.5-turbo \cite{gpt3.5} and GPT-3-Curie \cite{gpt3}. Based on the results of preliminary experiments, we augment 40\% of the outputs by random deletion, scrambling and swapping of the original and simplified sentences. As no annotated reference simplification is required, this stage is arbitrarily scalable, and the set of supervision signals is extendable. 

\subsection{Pretraining with Reference-Free and Reference-Based Supervision Signals}

As highlighted earlier, TS data with human ratings are scarce, which creates a bottleneck for TS evaluation. However, there exist several TS corpora of aligned complex and human-simplified sentences, and in the second stage, we utilize this data in the form of reference-based supervision signals to provide more accurate supervision. Specifically, the second pretraining stage includes BLEU, SARI and BERTScore as supervision signals in addition to the reference-free signals from Stage 1. We utilize the Newsela \citeplanguageresource{newsela}, WikiSmall and WikiLarge \citeplanguageresource{wiki} test sets, totalling at 1,536 (source, reference) pairs. In addition to the outputs from the models used in the previous stage, we include outputs from EditNTS \cite{editnts}, DRESS \cite{dress}, Hybrid \cite{hybrid}, and PBMT-R \cite{pbmt-r}. Due to the reliance on reference simplifications, this stage is scalable only in terms of the number of simplification systems. 

\subsection{Fine-tuning}

Finally, we fine-tune the metric on human ratings such that it is aligned with the particular criteria used in each dataset. We consider fine-tuning on overall as well as specific ratings. For overall ratings, we use the \textsc{SimpEval} \cite{lens} corpus which contains overall quality ratings from five annotators. Following \citet{lens}, we use the \textsc{SimpEval}$_{\mathrm{PAST}}$ and \textsc{SimpEval}$_{2022}$ subsets of the \textsc{SimpEval} corpus respectively for training and evaluation. \textsc{SimpEval}$_{\mathrm{PAST}}$ contains ratings for 2.4K system outputs from 24 systems and is based on a subset of the TurkCorpus \citeplanguageresource{turkcorpus}. \textsc{SimpEval}$_{2022}$ is designed to present a more challenging scenario and contains 360 simplifications from 6 systems for a new, curated set of sentences that are longer and discuss recent events. In particular, \textsc{SimpEval}$_{2022}$ includes only higher-quality simplifications from GPT-3.5~\cite{gpt3.5}, T5~\cite{t5}, MUSS~\cite{muss} and human annotators. We report the aggregated results from three runs.

To evaluate our method on learning specific scores, we utilize the \textsc{Simplicity-DA}~\citeplanguageresource{simplicity_da} and \textsc{Human-Likert}~\citeplanguageresource{scialom2021rethinking_data} datasets. Simplicity-DA contains 600 system outputs from six systems from the TurkCorpus~\citeplanguageresource{turkcorpus} test set, each with annotations on adequacy, fluency and simplicity. Human-Likert follows the same format and contains 112 human-written simplifications from the ASSET\citeplanguageresource{asset} and TurkCorpus test sets. The smaller size of the two datasets presents a challenge to learnable metrics. We fine-tune and evaluate \textsc{REFeREE} separately on the different aspects. We split the datasets into training, validation and test sets by source sentences with a 4-1-1 ratio and report the averaged results on five runs with non-overlapping test sets.

\begin{table*}[ht]
\centering
\resizebox{1.5\columnwidth}{!}{%
\begin{tabular}{lccc}
\hline
\textbf{Metric} & $\tau_{paraphrase} \uparrow$ & $\tau_{splitting} \uparrow$ & $\tau_{all} \uparrow$ \\
\hline
FKGL & $-0.556$ & $-0.310$ & $-0.356$ \\
BLEU & $-0.048$ & $-0.054$ & $-0.033$ \\
SARI & $0.397$ & $0.264$ & $0.289$ \\
BERTScore$_{\mathrm{F1}}$ & $0.175$ & $0.023$ & $0.052$ \\
BERTScore$_{\mathrm{precision}}$ & $0.238$ & $0.093$ & $0.112$ \\
self-BERTScore$_{\mathrm{F1}}$ & $-0.174$ & $-0.333$ & $-0.300$ \\
self-BERTScore$_{\mathrm{precision}}$ & $-0.079$ & $-0.348$ & $-0.300$ \\
\hline
BLEURT$_{\mathrm{WMT}}$ & $0.055$ & $0.073$ & $0.030$ \\
BLEURT$_{\mathrm{fine-tuned}}$ & $0.270 \pm 0.113$ & $0.132 \pm 0.023$ & $0.163 \pm 0.040$ \\
\textsc{BETS} & $-0.302$ & $-0.349$ & $-0.331$ \\
\textsc{SLE} & $0.492$ & $0.256$ & $0.295$ \\
\textsc{Lens}$_{k=3}$ & $0.429$ & $0.333$ & $0.331$ \\
\hline
\textsc{REFeREE} & $ 0.481 \pm 0.015$  & $ \mathbf{0.341 \pm 0.029} $ & $ 0.360 \pm 0.020 $ \\
\textsc{REFeREE} (RoBERTa) & $\mathbf{0.534 \pm 0.030}$ & $0.328 \pm 0.019$ & $\mathbf{0.368 \pm 0.018}$ \\
\hline
\end{tabular}
}
\caption{Results on the \textsc{SimpEval}$_{2022}$ dataset for different operation types. We use the official checkpoints for off-the-shelf metrics. For the metrics we trained (\textsc{REFeREE} and BLEURT$_{\mathrm{fine-tuned}}$), we report the aggregated results from three runs. We follow \citet{lens} and report the Kendall Tau-like coefficient on filtered pairs where all three annotators agree with the ranking order and the unnormalized score difference (out of 100) is larger than five for at least two annotators. Since each operation group is rated separately for the dataset, we do not compare simplifications of different operation types. For the same reason, the overall Pearson correlation is not compatible with this dataset.}
\label{tab:simpeval_results}
\end{table*}

\section{Experimental Results}

We fine-tune and evaluate \textsc{REFeREE} separately on the three datasets. For each stage, the model is trained on the unweighted average of L2 losses from the training signals. We use the Adam optimizer \cite{adam} with $\epsilon$ = $10^{-6}$, $\beta_1$ = $0.9$, and $\beta_2$ = $0.999$, learning rates of $10^{-5}$, $10^{-5}$, and $10^{-7}$ for the three stages, and perform early-stopping based on the development set performance. 

Human ratings are aggregated by taking the average. For \textsc{Simplicity-DA} and \textsc{Human-Likert}, we report the Pearson correlation $r$. For \textsc{SimpEval}, as the ratings for different simplification operations (deletion, paraphrase and splitting) are separately collected and not comparable, we follow \citet{lens} and report the Kendall Tau-like coefficient $\tau$ \cite{wmt17}, which has a range between -1 and 1 and is defined as:
$$
\tau = \frac{|Concordant| - |Discordant|}{|Concordant| + |Discordant|}
$$
where $Concordant$ is the set of pair-wise rankings where the metric agrees with human ratings on simplifications for the same source sentence, and $Discordant$ is the set of rankings where they disagree. 

We compare our metric with non-learnable metrics (FKGL, BLEU, SARI, self-BERTScore and BERTScore), as well as BLEURT \cite{bleurt_scaled_up}, \textsc{Lens}, BETS, and SLE.\footnote{We include the precision score of BERTScore in addition to the F1 score as \citetlanguageresource{simplicity_da} observe that it correlates better with human judgments. BLEURT was originally fine-tuned on WMT datasets \cite{wmt19}. We also experiment with fine-tuning it on the simplification datasets. We compare against the highest-performing \textsc{Lens} variant trained using the top-3 references and the highest-performing SLE variant measuring relative simplicity and trained on softened labels. In the cases where the simplicity branch of BETS fails to return a score, we assign a simplicity score of 0.} For a fair comparison with \textsc{Lens}, we also include a variant of \textsc{REFeREE} with a RoBERTa-large backbone. Finally, to investigate the contribution of the three training stages, data augmentation and training signals, we also report ablation results on \textsc{SimpEval}. 

\subsection{Overall Ratings}

Results on the \textsc{SimpEval}$_{2022}$ dataset are shown in Table \ref{tab:simpeval_results}.\footnote{We observe slightly different results than those reported in \citet{lens}, which is likely due to the use of different implementations of the evaluation metrics.} Non-learnable metrics perform poorly, with FKGL, BLEU and self-BERTScore having more discordant pairs than concordant ones, demonstrating the need for learnable metrics. In particular, self-BERTScore measures sentence similarity and may mistakenly punish simplifications that remove non-essential information or inadequately punish under-simplifications. The off-the-shelf and fine-tuned BLEURT metrics also perform poorly as the BLEURT pretraining process mainly considers semantic similarity and does not adequately encompass all aspects of simplification quality. Somewhat surprisingly, BETS performs very poorly for all types of simplification. We hypothesize that this is because it is primary designed for lexical simplification and cannot effectively evaluate simplification by syntactic changes. SLE, a metric trained to predict simplicity, performs relatively well on this dataset, suggesting that the overall quality ratings are correlated with sentence simplicity and that the simplifications in \textsc{SimpEval}$_{2022}$ have relatively few adequacy and fluency issues.

\textsc{REFeREE} outperforms \textsc{Lens}, which uses a larger model (354M parameters compared with 226M parameters) and relies on more information at inference time. Changing to the larger RoBERTa-large backbone results in slightly improved overall performance, with a significant improvement in evaluating paraphrases. A further inspection reveals that compared with \textsc{Lens}, \textsc{REFeREE} is more effective at evaluating machine simplifications and relatively underperforms when handling human simplifications. Specifically, \textsc{REFeREE} results in a KendallTau-like coefficient of 0.310 when evaluating sentence pairs involving human simplifications and 0.500 when evaluating other pairs whereas the results for \textsc{Lens} are 0.371 and 0.273, respectively. We suspect that this is because \textsc{REFeREE} is less exposed to human-written simplifications during pretraining compared with machine simplifications.

\begin{table*}[ht]
\centering
\resizebox{1.5\columnwidth}{!}{%
\begin{tabular}{lccc}
\hline
\multirow{2}{*}{\textbf{Metric}} & \multicolumn{1}{c}{\textbf{Adequacy}} & \multicolumn{1}{c}{\textbf{Fluency}} & \multicolumn{1}{c}{\textbf{Simplicity}} \\
                       & $r$$\uparrow$ & $r$$\uparrow$ & $r$$\uparrow$\\
                        \hline

\hline
FKGL & $0.064\pm 0.164$ & $0.083\pm 0.207$ & $0.099\pm 0.117$ \\
BLEU & $0.354\pm 0.153$ & $0.317\pm 0.144$ & $0.221\pm 0.145$ \\
SARI & $0.258\pm 0.066$ & $0.164\pm 0.077$ & $0.180\pm 0.094$  \\
BERTScore$_{\mathrm{F1}}$ & $0.569\pm 0.075$ & $0.462\pm 0.108$ & $0.362\pm 0.066$ \\
BERTScore$_{\mathrm{Precision}}$ & $0.513\pm 0.095$ & $0.480\pm 0.111$ & $0.426\pm 0.074$ \\
self-BERTScore$_{\mathrm{F1}}$ & $\mathbf{0.727\pm 0.044}$ & $0.528\pm 0.083$ & $0.390\pm 0.045$ \\
self-BERTScore$_{\mathrm{Precision}}$ & $0.687\pm 0.069$ & $0.566\pm 0.054$ & $0.481\pm 0.041$ \\
\hline
BLEURT$_{\mathrm{WMT}}$ & $0.595\pm 0.082$ & $0.437\pm 0.172$ & $0.323\pm 0.111$ \\
BLEURT$_{\mathrm{fine-tuned}}$ & $0.096\pm 0.229$ & $0.384\pm 0.158$ & $0.150\pm 0.228$ \\
\textsc{BETS} & $0.592\pm 0.050$ & $0.367\pm 0.094$ & $0.155\pm 0.065$ \\
\textsc{SLE} & $-0.329\pm 0.097$ & $-0.128\pm 0.155$ & $0.018\pm 0.082$ \\
\textsc{Lens}$_{k=3}$ & $0.636\pm 0.069$ & $\mathbf{0.758} \pm \mathbf{0.059}$ & $\mathbf{0.732} \pm \mathbf{0.094}$ \\
\hline
\textsc{REFeREE} & $0.622\pm 0.079$  & $0.478 \pm 0.045$ & $0.366 \pm 0.126$ \\
% \textsc{REFeREE} (w/o stage 1) & $0.646\pm0.045$ & $0.619\pm 0.054$ & $0.536\pm 0.029$ \\
\textsc{REFeREE} (RoBERTa) & $0.633 \pm 0.038$ & $0.483\pm 0.57$ & $0.427\pm 0.058$ \\
\hline
\end{tabular}
}
\caption{Results on the \textsc{Simplicity-DA} dataset. The \textsc{Lens} model is trained on \textsc{SimpEval}$_{\mathrm{PAST}}$ and not fine-tuned on this dataset. The dataset is not compatible with the Kendall Tau-like coefficient as it mostly does not contain simplifications from different systems for the same source sentence.}
\label{tab: simplicity_da_results}
\end{table*}

\begin{table*}[ht]
\centering
\resizebox{1.5\columnwidth}{!}{%
\begin{tabular}{lccc}
\hline
\multirow{2}{*}{\textbf{Metric}} & \multicolumn{1}{c}{\textbf{Adequacy}} & \multicolumn{1}{c}{\textbf{Fluency}} & \multicolumn{1}{c}{\textbf{Simplicity}} \\
                       & $r$$\uparrow$ & $r$$\uparrow$ & $r$$\uparrow$\\
                        \hline

\hline
FKGL & $0.111 \pm 0.125$ & $-0.169\pm 0.136$ & $-0.385\pm 0.226$ \\
BLEU & $0.280 \pm 0.163$ & $0.316 \pm 0.183$ & $0.157 \pm 0.209$ \\
SARI & $0.139 \pm 0.120$ & $0.236 \pm 0.089$ & $0.445 \pm 0.108$ \\
BERTScore$_{\mathrm{F1}}$ & $0.280 \pm 0.150$ & $0.214 \pm 0.042$ & $0.105 \pm 0.130$ \\
BERTScore$_{\mathrm{Precision}}$ & $0.266 \pm 0.192$ & $0.433 \pm 0.065$ & $0.321 \pm 0.077$ \\
self-BERTScore$_{\mathrm{F1}}$ & $0.421 \pm 0.169$ & $0.016 \pm 0.103$ & $-0.385 \pm 0.108$ \\
self-BERTScore$_{\mathrm{Precision}}$ & $0.345 \pm 0.210$ & $0.175 \pm 0.107$ & $-0.255 \pm 0.128$ \\
\hline
BLEURT$_{\mathrm{WMT}}$ & $0.441 \pm 0.047$ & $0.247 \pm 0.022$ & $0.093 \pm 0.072$ \\
BLEURT$_{\mathrm{fine-tuned}}$ & $\mathbf{0.472\pm 0.110}$ & $0.251\pm 0.150$ & $0.077\pm 0.109$ \\
\textsc{BETS} & $0.375 \pm 0.167$ & $-0.114 \pm 0.106$ & $-0.513 \pm 0.066$ \\
\textsc{SLE} & $-0.206 \pm 0.142$ & $0.186 \pm 0.099$ & $0.532 \pm 0.072$ \\
\textsc{Lens}$_{k=3}$ & $0.201 \pm 0.122$ & $\mathbf{0.561 \pm 0.057}$ & $\mathbf{0.561 \pm 0.055}$ \\
\hline
\textsc{REFeREE} & $0.425 \pm 0.127$ & $0.308\pm 0.108$ & $0.322\pm 0.075$ \\
\textsc{REFeREE} (RoBERTa) & $0.386\pm 0.120$ & $0.292\pm 0.171$ & $0.528\pm 0.083$ \\
\hline
\end{tabular}
}
\caption{Results on the \textsc{Human-Likert} dataset. The \textsc{Lens} model is trained on \textsc{SimpEval}$_{\mathrm{PAST}}$ and not fine-tuned on this dataset. The dataset is not compatible with the Kendall Tau-like coefficient as it mostly does not contain simplifications from different systems for the same source sentence.}
\label{tab: human_likert_results}
\end{table*}

\subsection{Specific Ratings}
Tables \ref{tab: simplicity_da_results} and \ref{tab: human_likert_results} show the results on the \textsc{Simplicity-DA} and \textsc{Human-Likert} datasets for adequacy, fluency and simplicity. Overall, \textsc{REFeREE} underperforms \textsc{Lens} but still performs better than conventional metrics. On \textsc{Simplicity-DA}, we observe that most source-based metrics (self-BERTScore, \textsc{Lens} and \textsc{REFeREE}) outperform source-free metrics, suggesting that having direct access to the source sentence helps with measuring meaning preservation. Self-BERTScore, with the highest performance on adequacy, also performs well in measuring fluency and simplicity. This is likely due to the high intra-correlation between the three aspects in Simplicity-DA (e.g. inadequate simplifications are likely not fluent and difficult to understand) as pointed out by \cite{scialom2021rethinking}. \textsc{SLE}, despite performing well on \textsc{Simplicity-DA} when controlling for adequacy and fluency \cite{sle}, performs poorly on the unfiltered dataset as it is not exposed to the lower-quality machine simplifications during its training process. The relative underperformance of \textsc{REFeREE} is likely because the dataset includes outputs from several dated systems absent in our pretraining process. Incorporating more systems and refining the augmentation process will likely lead to improvements.

We observe that the performance of most metrics is inconsistent between \textsc{Simplicity-DA} and \textsc{Human-Likert}. For instance, \textsc{Lens} performs much worse in predicting adequacy scores on \textsc{Human-Likert} than on \textsc{Simplicity-DA}. Aside from \textsc{Human-Likert} having lower inter-correlations, this can also be due to the variances in the collection of human ratings (e.g. annotators and criteria). Despite having lower overall performance, \textsc{REFeREE} performs more consistently between the datasets as it can be fine-tuned to be aligned with the particularities of each set of ratings.

% \textsc{REFeREE} appears to be effective with the small datasets, reaching comparable overall performance with \textsc{Lens}. In particular, \textsc{REFeREE} constistently performs better in predicting adequacy as it has direct access to the original sentence and is fine-tuned on the human ratings. On \textsc{Simplicity-DA}, we observe that the pretraining stages lead to worsened performance in fluency and simplicity, which is likely because the dataset includes outputs from several dated systems absent in our pretraining process. Incorporating more systems and refining the augmentation process will likely lead to improvements.

\begin{table*}[ht]
\centering
\resizebox{1.4\columnwidth}{!}{%
\begin{tabular}{lccc}
\hline
\textbf{Metric} & $\tau_{paraphrase} \uparrow$ & $\tau_{splitting} \uparrow$ & $\tau_{all} \uparrow$ \\
\hline
\textsc{REFeREE} & $\mathbf{ 0.481 \pm 0.015}$ & $ 0.341 \pm 0.029 $ & $ 0.360 \pm 0.020 $ \\
\hline
Stage 1 + fine-tuning & $0.439 \pm 0.054$ & $0.370 \pm 0.041$ & $\mathbf{0.374 \pm 0.031}$ \\
Stage 2 + fine-tuning &  $ 0.365 \pm 0.052 $  &  $ 0.240 \pm 0.058 $  &  $ 0.254 \pm 0.058 $  \\
Fine-tuning only & $ 0.354 \pm 0.030 $ & $ 0.220 \pm 0.045 $ & $ 0.242 \pm 0.037 $ \\
\hline
\textsc{REFeREE}, all stages &&&\\
w/o fluency &   $0.376 \pm 0.015$ & $0.328 \pm 0.007$ & $0.335 \pm 0.003$ \\
w/o meaning &   $0.407 \pm 0.030$ & $\mathbf{0.372 \pm 0.013}$ & $0.358 \pm 0.006$ \\
w/o simplicity &   $0.418 \pm 0.015$ & $0.305 \pm 0.015$ & $0.309 \pm 0.016$ \\
w/o augmentation &   $0.439 \pm 0.015$ & $0.271 \pm 0.006$ & $0.289 \pm 0.009$ \\
\hline
\end{tabular}
}
\caption{Ablation results on \textsc{SimpEval}$_{2022}$.}
% \caption{Ablation results on \textsc{SimpEval}$_{2022}$. ($*$) indicates statistically lower correlation than the unablated model ($p < 0.05$) based on a percentile bootstrap with 10K resamples.}
\label{tab:ablation_results}
\end{table*}

\subsection{Ablation Experiments}
Finally, to shed light on how the different components of \textsc{REFeREE} affect the overall performance, we report the ablation study results on \textsc{SimpEval}$_{2022}$ with respect to the training regime, the types of supervision signals and data augmentation. Comparing variants of \textsc{REFeREE} with different training regimes, we find that the first pretraining stage significantly improves the performance thanks to its relatively large scale, leading to an increase of over 0.13 in the Kendall Tau-like coefficient for all operation types compared with fine-tuning only. The second pretraining stage appears limitedly helpful when compared with fine-tuning only and even results in slightly degraded performance when combined with the first pretraining stage. This can be due to potentially varying quality of the references and the small dataset size for this stage, which may more easily lead to overfitting. This further signifies the utility of the arbitrarily scalable first pretraining stage as a means to improve model performance under the scarcity of human-annotated ratings.

We also experiment with ablating types of supervision signals for meaning preservation (BLEU, self-BLEU, BERTScore and self-BERTScore), fluency (GPT-2 perplexity) and simplicity (FKGL and CommonLit readability). We observe that the supervision signals for meaning preservation only slightly influence the performance, which is likely because the machine simplifications in SimpEval2022 are of high quality and generally preserve the meaning of the complex sentences. The signals for fluency and simplicity seem to play a more impactful role, likely because the systems in SimpEval2022 (GPT-3.5, T5, MUSS and humans) produce simplifications in different styles and of varying fluency. This explains why our metric outperforms non-specialized metrics such as BERTScore which primarily focus on meaning preservation.

Ablating data augmentation leads to a significant decrease in the metric’s performance. This is because the first pre-training stage involves relatively high-quality simplification produced by MUSS and GPT, and data augmentation is an effective way of generating lower-quality simplifications that complement these system outputs.

\section{Discussion}

Despite promising results demonstrated in this work, we also recognize that there are limitations of the proposed method. In this section, we discuss such limitations, outlining potential directions for future research.

First of all, due to the data limitations, \textsc{REFeREE} is only fine-tuned and evaluated on sentence-level simplifications in the English news and Wikipedia domains. Its performance on other languages, domains and simplification setups (e.g. document-level simplification \cite{doc_level_simplification} and elaborative simplification \cite{elaborative_simplification}) awaits further investigation. Whereas our reference-free pretraining stage is arbitrarily scalable in design, we only experimented with a reasonably small pretraining dataset of around 200K source sentences simplified by three systems. Further experiments are required to determine how the performance of the metric scales with the dataset size and the number of simplification systems. In addition, since the metric is fine-tuned on small datasets, its out-of-domain performance on other datasets and simplification systems is not guaranteed.

% Finally, despite the increasing need in reference-free evaluation metrics and the proposal of several approaches to such metrics development, applicability of such metrics needs to be carefully considered. \citet{deutsch-etal-2022-limitations} discussed the limitations of reference-free evaluations of generated text, highlighting that ``reference-free metrics are equivalent to using one generation model to evaluate another", which means that the metrics can be optimized at test time to find the approximate best-possible output, they are inherently biased toward models which are more similar to their own, and they can be biased against higher-quality outputs, including those written by humans. Instead of using reference-free metrics to evaluate models on the specific tasks, \citet{deutsch-etal-2022-limitations} recommended that they should be used as diagnostic tools for analyzing and understanding model behavior. This is explored through an analysis of three reference-free evaluation metrics: Prism-src (Thompson and Post, 2020) and COMET-QE~\cite{rei-etal-2021-references} for MT and QuestEval~\cite{scialom-etal-2021-questeval} for summarization.
% At the same time,~\citet{louis2013automatically} highlight that reference-free metrics they proposed are used to ``complement but are not intended to replace existing manual and automatic approaches to evaluation wherein the latter’s strength and reliability are important for high confidence evaluations".

Finally, despite the increasing need for reference-free evaluation metrics and the development of multiple reference-free approaches, the applicability of such metrics needs to be carefully considered. \citet{deutsch-etal-2022-limitations} highlight that the metrics can be over-optimized at test time, and may be biased both towards models similar to their backbones and against higher-quality outputs produced by humans. We agree with~\citet{deutsch-etal-2022-limitations} that reference-free metrics should be used as diagnostic tools and with \citet{louis2013automatically} that these metrics should complement high-quality human evaluation. However, we also contend that this still makes them useful during the rapid prototyping of new systems where human evaluations are costly, difficult or sometimes impossible to collect.

% TODO: Include and discuss limitations of the study (which may be also phrased as future work):

% \begin{itemize}
%     \item Trainable / learnable metrics require additional resources and time for training, whereas other (i.e., traditional) metrics are light-weight. How computationally expensive is it to train / apply REFeREE?
%     \item So far, this work addresses evaluation with respect to the quality of the generated output. We leave the second type of evaluation (testing with human subjects /  target population) as future work.
%     \item We should address limitations of learnable metrics from \citet{deutsch-etal-2022-limitations}. Do they apply to this work?
%     \item Can we say anything about generalization behavior of the metric? Does it need retraining on any new data? How does it deal with the subjectivity of human annotation in TS? 
% \end{itemize}

\section{Conclusion}

We propose \textsc{REFeREE}, a reference-free model-based metric for text simplification with a 3-stage curriculum, including an arbitrarily scalable pretraining using reference-free supervision signals as well as pretraining with both reference-free and reference-based supervision signals, and a fine-tuning stage with human ratings. Our experiments show that our metric is effective and flexible, attaining competitive performance in evaluating with respect to both general and specific ratings of the quality of the text simplification system outputs.

Since the formulation of our metric is largely generalizable, it can be modified and applied to other conditional natural language generation tasks such as abstractive summarization, among others. This calls for an investigation into task-agnostic and multi-task supervision, which we leave as future work.

\section*{Acknowledgements}

This work is financially supported by Mohamed bin Zayed University of Artificial Intelligence (MBZUAI) and is supported by the Campus Super Computing Center at MBZUAI. We thank the anonymous reviewers for their valuable feedback.

\nocite{*}
\section{Bibliographical References}\label{sec:reference}

\bibliographystyle{lrec-coling2024-natbib}
\bibliography{bibliographical_references}

\begin{thebibliography}{8}
\expandafter\ifx\csname natexlab\endcsname\relax\def\natexlab#1{#1}\fi

\bibitem[{Alva-Manchego et~al.(2020)Alva-Manchego, Martin, Bordes, Scarton,
  Sagot, and Specia}]{asset}
Fernando Alva-Manchego, Louis Martin, Antoine Bordes, Carolina Scarton,
  Beno{\^\i}t Sagot, and Lucia Specia. 2020.
\newblock \href {https://www.aclweb.org/anthology/2020.acl-main.424} {{ASSET}:
  {A} dataset for tuning and evaluation of sentence simplification models with
  multiple rewriting transformations}.
\newblock In \emph{Proceedings of the 58th Annual Meeting of the Association
  for Computational Linguistics}, pages 4668--4679, Online. Association for
  Computational Linguistics.

\bibitem[{Alva-Manchego et~al.(2021)Alva-Manchego, Scarton, and
  Specia}]{simplicity_da}
Fernando Alva-Manchego, Carolina Scarton, and Lucia Specia. 2021.
\newblock \href {https://doi.org/10.1162/coli_a_00418} {{The (Un)Suitability of
  Automatic Evaluation Metrics for Text Simplification}}.
\newblock \emph{Computational Linguistics}, 47(4):861--889.

\bibitem[{Gokaslan and Cohen(2019)}]{openwebtext}
Aaron Gokaslan and Vanya Cohen. 2019.
\newblock \href {http://Skylion007.github.io/OpenWebTextCorpus} {{OpenWebText
  Corpus}}.

\bibitem[{Maddela et~al.(2023)Maddela, Dou, Heineman, and Xu}]{lens}
Mounica Maddela, Yao Dou, David Heineman, and Wei Xu. 2023.
\newblock \href {https://doi.org/10.18653/v1/2023.acl-long.905} {{LENS}: A
  learnable evaluation metric for text simplification}.
\newblock In \emph{Proceedings of the 61st Annual Meeting of the Association
  for Computational Linguistics (Volume 1: Long Papers)}, pages 16383--16408,
  Toronto, Canada. Association for Computational Linguistics.

\bibitem[{Scialom et~al.(2021)Scialom, Martin, Staiano, de~La~Clergerie, and
  Sagot}]{scialom2021rethinking_data}
Thomas Scialom, Louis Martin, Jacopo Staiano, Eric~Villemonte de~La~Clergerie,
  and Beno{\^\i}t Sagot. 2021.
\newblock {Rethinking automatic evaluation in sentence simplification}.
\newblock \emph{arXiv preprint arXiv:2104.07560}.

\bibitem[{Xu et~al.(2015)Xu, Callison-Burch, and Napoles}]{newsela}
Wei Xu, Chris Callison-Burch, and Courtney Napoles. 2015.
\newblock \href {https://doi.org/10.1162/tacl_a_00139} {{Problems in Current
  Text Simplification Research: New Data Can Help}}.
\newblock \emph{Transactions of the Association for Computational Linguistics},
  3:283--297.

\bibitem[{Xu et~al.(2016)Xu, Napoles, Pavlick, Chen, and
  Callison-Burch}]{turkcorpus}
Wei Xu, Courtney Napoles, Ellie Pavlick, Quanze Chen, and Chris Callison-Burch.
  2016.
\newblock \href {https://doi.org/10.1162/tacl_a_00107} {{Optimizing Statistical
  Machine Translation for Text Simplification}}.
\newblock \emph{Transactions of the Association for Computational Linguistics},
  4:401--415.

\bibitem[{Zhang and Lapata(2017)}]{wiki}
Xingxing Zhang and Mirella Lapata. 2017.
\newblock \href {http://aclweb.org/anthology/D17-1063} {{Sentence
  Simplification with Deep Reinforcement Learning}}.
\newblock In \emph{Proceedings of the 2017 Conference on Empirical Methods in
  Natural Language Processing}, pages 595--605. Association for Computational
  Linguistics.

\end{thebibliography}


\begin{thebibliography}{62}
\expandafter\ifx\csname natexlab\endcsname\relax\def\natexlab#1{#1}\fi

\bibitem[{Alva-Manchego et~al.(2019)Alva-Manchego, Martin, Scarton, and
  Specia}]{easse}
Fernando Alva-Manchego, Louis Martin, Carolina Scarton, and Lucia Specia. 2019.
\newblock \href {https://doi.org/10.18653/v1/D19-3009} {{{EASSE}: Easier
  Automatic Sentence Simplification Evaluation}}.
\newblock In \emph{Proceedings of the 2019 Conference on Empirical Methods in
  Natural Language Processing and the 9th International Joint Conference on
  Natural Language Processing (EMNLP-IJCNLP): System Demonstrations}, pages
  49--54, Hong Kong, China. Association for Computational Linguistics.

\bibitem[{Alva-Manchego et~al.(2021)Alva-Manchego, Scarton, and
  Specia}]{simplicity_da_eval}
Fernando Alva-Manchego, Carolina Scarton, and Lucia Specia. 2021.
\newblock \href {https://doi.org/10.1162/coli_a_00418} {{The (Un)Suitability of
  Automatic Evaluation Metrics for Text Simplification}}.
\newblock \emph{Computational Linguistics}, 47(4):861--889.

\bibitem[{Barrault et~al.(2019)Barrault, Bojar, Costa-juss{\`a}, Federmann,
  Fishel, Graham, Haddow, Huck, Koehn, Malmasi, Monz, M{\"u}ller, Pal, Post,
  and Zampieri}]{wmt19}
Lo{\"\i}c Barrault, Ond{\v{r}}ej Bojar, Marta~R. Costa-juss{\`a}, Christian
  Federmann, Mark Fishel, Yvette Graham, Barry Haddow, Matthias Huck, Philipp
  Koehn, Shervin Malmasi, Christof Monz, Mathias M{\"u}ller, Santanu Pal, Matt
  Post, and Marcos Zampieri. 2019.
\newblock \href {https://doi.org/10.18653/v1/W19-5301} {{Findings of the 2019
  Conference on Machine Translation ({WMT}19)}}.
\newblock In \emph{Proceedings of the Fourth Conference on Machine Translation
  (Volume 2: Shared Task Papers, Day 1)}, pages 1--61, Florence, Italy.
  Association for Computational Linguistics.

\bibitem[{Belouadi and Eger(2023)}]{uscore}
Jonas Belouadi and Steffen Eger. 2023.
\newblock \href {https://aclanthology.org/2023.eacl-main.27} {{{US}core: An
  Effective Approach to Fully Unsupervised Evaluation Metrics for Machine
  Translation}}.
\newblock In \emph{Proceedings of the 17th Conference of the European Chapter
  of the Association for Computational Linguistics}, pages 358--374, Dubrovnik,
  Croatia. Association for Computational Linguistics.

\bibitem[{Bojar et~al.(2017)Bojar, Graham, and Kamran}]{wmt17}
Ond{\v{r}}ej Bojar, Yvette Graham, and Amir Kamran. 2017.
\newblock \href {https://doi.org/10.18653/v1/W17-4755} {{Results of the {WMT}17
  Metrics Shared Task}}.
\newblock In \emph{Proceedings of the Second Conference on Machine
  Translation}, pages 489--513, Copenhagen, Denmark. Association for
  Computational Linguistics.

\bibitem[{Brown et~al.(2020)Brown, Mann, Ryder, Subbiah, Kaplan, Dhariwal,
  Neelakantan, Shyam, Sastry, Askell, Agarwal, Herbert-Voss, Krueger, Henighan,
  Child, Ramesh, Ziegler, Wu, Winter, Hesse, Chen, Sigler, Litwin, Gray, Chess,
  Clark, Berner, McCandlish, Radford, Sutskever, and Amodei}]{gpt3}
Tom Brown, Benjamin Mann, Nick Ryder, Melanie Subbiah, Jared~D Kaplan, Prafulla
  Dhariwal, Arvind Neelakantan, Pranav Shyam, Girish Sastry, Amanda Askell,
  Sandhini Agarwal, Ariel Herbert-Voss, Gretchen Krueger, Tom Henighan, Rewon
  Child, Aditya Ramesh, Daniel Ziegler, Jeffrey Wu, Clemens Winter, Chris
  Hesse, Mark Chen, Eric Sigler, Mateusz Litwin, Scott Gray, Benjamin Chess,
  Jack Clark, Christopher Berner, Sam McCandlish, Alec Radford, Ilya Sutskever,
  and Dario Amodei. 2020.
\newblock \href
  {https://proceedings.neurips.cc/paper_files/paper/2020/file/1457c0d6bfcb4967418bfb8ac142f64a-Paper.pdf}
  {{Language Models are Few-Shot Learners}}.
\newblock In \emph{Advances in Neural Information Processing Systems},
  volume~33, pages 1877--1901. Curran Associates, Inc.

\bibitem[{Chandrasekar and Srinivas(1997)}]{chandrasekar1997automatic}
Raman Chandrasekar and Bangalore Srinivas. 1997.
\newblock Automatic induction of rules for text simplification.
\newblock \emph{Knowledge-Based Systems}, 10(3):183--190.

\bibitem[{Cripwell et~al.(2023)Cripwell, Legrand, and Gardent}]{sle}
Liam Cripwell, Jo{\"e}l Legrand, and Claire Gardent. 2023.
\newblock {Simplicity Level Estimate (SLE): A Learned Reference-Less Metric for
  Sentence Simplification}.
\newblock \emph{arXiv preprint arXiv:2310.08170}.

\bibitem[{Deutsch et~al.(2022)Deutsch, Dror, and
  Roth}]{deutsch-etal-2022-limitations}
Daniel Deutsch, Rotem Dror, and Dan Roth. 2022.
\newblock \href {https://doi.org/10.18653/v1/2022.emnlp-main.753} {{On the
  Limitations of Reference-Free Evaluations of Generated Text}}.
\newblock In \emph{Proceedings of the 2022 Conference on Empirical Methods in
  Natural Language Processing}, pages 10960--10977, Abu Dhabi, United Arab
  Emirates. Association for Computational Linguistics.

\bibitem[{Dong et~al.(2019)Dong, Li, Rezagholizadeh, and Cheung}]{editnts}
Yue Dong, Zichao Li, Mehdi Rezagholizadeh, and Jackie Chi~Kit Cheung. 2019.
\newblock \href {https://doi.org/10.18653/v1/P19-1331} {{{E}dit{NTS}: An Neural
  Programmer-Interpreter Model for Sentence Simplification through Explicit
  Editing}}.
\newblock In \emph{Proceedings of the 57th Annual Meeting of the Association
  for Computational Linguistics}, pages 3393--3402, Florence, Italy.
  Association for Computational Linguistics.

\bibitem[{Fonseca et~al.(2019)Fonseca, Yankovskaya, Martins, Fishel, and
  Federmann}]{fonseca-etal-2019-findings}
Erick Fonseca, Lisa Yankovskaya, Andr{\'e} F.~T. Martins, Mark Fishel, and
  Christian Federmann. 2019.
\newblock \href {https://doi.org/10.18653/v1/W19-5401} {{Findings of the {WMT}
  2019 Shared Tasks on Quality Estimation}}.
\newblock In \emph{Proceedings of the Fourth Conference on Machine Translation
  (Volume 3: Shared Task Papers, Day 2)}, pages 1--10, Florence, Italy.
  Association for Computational Linguistics.

\bibitem[{Fu et~al.(2023)Fu, Ng, Jiang, and Liu}]{gptscore}
Jinlan Fu, See-Kiong Ng, Zhengbao Jiang, and Pengfei Liu. 2023.
\newblock \href {http://arxiv.org/abs/2302.04166} {Gptscore: Evaluate as you
  desire}.

\bibitem[{Harman and Over(2004)}]{harman-over-2004-effects}
Donna Harman and Paul Over. 2004.
\newblock \href {https://aclanthology.org/W04-1003} {{The Effects of Human
  Variation in {DUC} Summarization Evaluation}}.
\newblock In \emph{Text Summarization Branches Out}, pages 10--17, Barcelona,
  Spain. Association for Computational Linguistics.

\bibitem[{He et~al.(2023)He, Gao, and Chen}]{debertav3}
Pengcheng He, Jianfeng Gao, and Weizhu Chen. 2023.
\newblock \href {https://openreview.net/forum?id=sE7-XhLxHA} {De{BERT}av3:
  Improving de{BERT}a using {ELECTRA}-style pre-training with
  gradient-disentangled embedding sharing}.
\newblock In \emph{The Eleventh International Conference on Learning
  Representations}.

\bibitem[{Hessel et~al.(2021)Hessel, Holtzman, Forbes, Le~Bras, and
  Choi}]{clip_score}
Jack Hessel, Ari Holtzman, Maxwell Forbes, Ronan Le~Bras, and Yejin Choi. 2021.
\newblock \href {https://doi.org/10.18653/v1/2021.emnlp-main.595}
  {{{CLIPS}core: A Reference-free Evaluation Metric for Image Captioning}}.
\newblock In \emph{Proceedings of the 2021 Conference on Empirical Methods in
  Natural Language Processing}, pages 7514--7528, Online and Punta Cana,
  Dominican Republic. Association for Computational Linguistics.

\bibitem[{Kajiwara and Fujita(2017)}]{kajiwara-fujita-2017-semantic}
Tomoyuki Kajiwara and Atsushi Fujita. 2017.
\newblock \href {https://aclanthology.org/I17-2019} {{Semantic Features Based
  on Word Alignments for Estimating Quality of Text Simplification}}.
\newblock In \emph{Proceedings of the Eighth International Joint Conference on
  Natural Language Processing (Volume 2: Short Papers)}, pages 109--115,
  Taipei, Taiwan. Asian Federation of Natural Language Processing.

\bibitem[{Kincaid et~al.(1975)Kincaid, Fishburne, Rogers, and Chissom}]{fkgl}
J.~Peter Kincaid, Robert~P. Fishburne, Richard~L. Rogers, and Brad~S. Chissom.
  1975.
\newblock {Derivation of New Readability Formulas (Automated Readability Index,
  Fog Count and Flesch Reading Ease Formula) for Navy Enlisted Personnel}.
\newblock Technical report, Naval Technical Training Command Millington TN
  Research Branch.

\bibitem[{Kingma and Ba(2015)}]{adam}
Diederik~P. Kingma and Jimmy Ba. 2015.
\newblock \href {http://dblp.uni-trier.de/db/conf/iclr/iclr2015.html#KingmaB14}
  {{Adam: A Method for Stochastic Optimization.}}
\newblock In \emph{ICLR (Poster)}.

\bibitem[{Kriz et~al.(2020)Kriz, Apidianaki, and
  Callison-Burch}]{kriz2020simple}
Reno Kriz, Marianna Apidianaki, and Chris Callison-Burch. 2020.
\newblock {Simple-QE: Better Automatic Quality Estimation for Text
  Simplification}.
\newblock \emph{arXiv preprint arXiv:2012.12382}.

\bibitem[{Lan et~al.(2020)Lan, Chen, Goodman, Gimpel, Sharma, and
  Soricut}]{albert}
Zhenzhong Lan, Mingda Chen, Sebastian Goodman, Kevin Gimpel, Piyush Sharma, and
  Radu Soricut. 2020.
\newblock \href {https://openreview.net/forum?id=H1eA7AEtvS} {{ALBERT: A Lite
  BERT for Self-supervised Learning of Language Representations}}.
\newblock In \emph{International Conference on Learning Representations}.

\bibitem[{Lin(2004)}]{rouge}
Chin-Yew Lin. 2004.
\newblock \href {https://aclanthology.org/W04-1013} {{{ROUGE}: A Package for
  Automatic Evaluation of Summaries}}.
\newblock In \emph{Text Summarization Branches Out}, pages 74--81, Barcelona,
  Spain. Association for Computational Linguistics.

\bibitem[{Liu et~al.(2023)Liu, Iter, Xu, Wang, Xu, and Zhu}]{g_eval}
Yang Liu, Dan Iter, Yichong Xu, Shuohang Wang, Ruochen Xu, and Chenguang Zhu.
  2023.
\newblock \href {http://arxiv.org/abs/2303.16634} {G-eval: Nlg evaluation using
  gpt-4 with better human alignment}.

\bibitem[{Liu et~al.(2019)Liu, Ott, Goyal, Du, Joshi, Chen, Levy, Lewis,
  Zettlemoyer, and Stoyanov}]{roberta}
Yinhan Liu, Myle Ott, Naman Goyal, Jingfei Du, Mandar Joshi, Danqi Chen, Omer
  Levy, Mike Lewis, Luke Zettlemoyer, and Veselin Stoyanov. 2019.
\newblock \href {http://arxiv.org/abs/1907.11692} {{RoBERTa: A Robustly
  Optimized BERT Pretraining Approach}}.

\bibitem[{Louis and Nenkova(2013)}]{louis2013automatically}
Annie Louis and Ani Nenkova. 2013.
\newblock {Automatically assessing machine summary content without a gold
  standard}.
\newblock \emph{Computational Linguistics}, 39(2):267--300.

\bibitem[{Maddela et~al.(2023)Maddela, Dou, Heineman, and Xu}]{lens}
Mounica Maddela, Yao Dou, David Heineman, and Wei Xu. 2023.
\newblock \href {https://doi.org/10.18653/v1/2023.acl-long.905} {{LENS}: A
  learnable evaluation metric for text simplification}.
\newblock In \emph{Proceedings of the 61st Annual Meeting of the Association
  for Computational Linguistics (Volume 1: Long Papers)}, pages 16383--16408,
  Toronto, Canada. Association for Computational Linguistics.

\bibitem[{Mallinson et~al.(2020)Mallinson, Severyn, Malmi, and Garrido}]{felix}
Jonathan Mallinson, Aliaksei Severyn, Eric Malmi, and Guillermo Garrido. 2020.
\newblock \href {https://doi.org/10.18653/v1/2020.findings-emnlp.111}
  {{{FELIX}: Flexible Text Editing Through Tagging and Insertion}}.
\newblock In \emph{Findings of the Association for Computational Linguistics:
  EMNLP 2020}, pages 1244--1255, Online. Association for Computational
  Linguistics.

\bibitem[{Martin et~al.(2022)Martin, Fan, de~la Clergerie, Bordes, and
  Sagot}]{muss}
Louis Martin, Angela Fan, {\'E}ric de~la Clergerie, Antoine Bordes, and
  Beno{\^\i}t Sagot. 2022.
\newblock \href {https://aclanthology.org/2022.lrec-1.176} {{{MUSS}:
  Multilingual Unsupervised Sentence Simplification by Mining Paraphrases}}.
\newblock In \emph{Proceedings of the Thirteenth Language Resources and
  Evaluation Conference}, pages 1651--1664, Marseille, France. European
  Language Resources Association.

\bibitem[{Martin et~al.(2018)Martin, Humeau, Mazar{\'e}, de~La~Clergerie,
  Bordes, and Sagot}]{martin-etal-2018-reference}
Louis Martin, Samuel Humeau, Pierre-Emmanuel Mazar{\'e}, {\'E}ric
  de~La~Clergerie, Antoine Bordes, and Beno{\^\i}t Sagot. 2018.
\newblock \href {https://doi.org/10.18653/v1/W18-7005} {{Reference-less Quality
  Estimation of Text Simplification Systems}}.
\newblock In \emph{Proceedings of the 1st Workshop on Automatic Text Adaptation
  ({ATA})}, pages 29--38, Tilburg, the Netherlands. Association for
  Computational Linguistics.

\bibitem[{Narayan and Gardent(2014)}]{hybrid}
Shashi Narayan and Claire Gardent. 2014.
\newblock \href {https://doi.org/10.3115/v1/P14-1041} {{Hybrid Simplification
  using Deep Semantics and Machine Translation}}.
\newblock In \emph{Proceedings of the 52nd Annual Meeting of the Association
  for Computational Linguistics (Volume 1: Long Papers)}, pages 435--445,
  Baltimore, Maryland. Association for Computational Linguistics.

\bibitem[{OpenAI(2023)}]{gpt3.5}
OpenAI. 2023.
\newblock \href {https://openai.com/blog/chatgpt/} {{ChatGPT: Optimizing
  language models for dialogue}}.

\bibitem[{Papineni et~al.(2002)Papineni, Roukos, Ward, and Zhu}]{bleu}
Kishore Papineni, Salim Roukos, Todd Ward, and Wei-Jing Zhu. 2002.
\newblock \href {https://doi.org/10.3115/1073083.1073135} {{{B}leu: a Method
  for Automatic Evaluation of Machine Translation}}.
\newblock In \emph{Proceedings of the 40th Annual Meeting of the Association
  for Computational Linguistics}, pages 311--318, Philadelphia, Pennsylvania,
  USA. Association for Computational Linguistics.

\bibitem[{Pu et~al.(2021)Pu, Chung, Parikh, Gehrmann, and
  Sellam}]{bleurt_scaled_up}
Amy Pu, Hyung~Won Chung, Ankur Parikh, Sebastian Gehrmann, and Thibault Sellam.
  2021.
\newblock \href {https://doi.org/10.18653/v1/2021.emnlp-main.58} {{Learning
  Compact Metrics for {MT}}}.
\newblock In \emph{Proceedings of the 2021 Conference on Empirical Methods in
  Natural Language Processing}, pages 751--762, Online and Punta Cana,
  Dominican Republic. Association for Computational Linguistics.

\bibitem[{Radford et~al.(2019)Radford, Wu, Child, Luan, Amodei, Sutskever
  et~al.}]{gpt2}
Alec Radford, Jeffrey Wu, Rewon Child, David Luan, Dario Amodei, Ilya
  Sutskever, et~al. 2019.
\newblock {Language Models are Unsupervised Multitask Learners}.
\newblock \emph{OpenAI blog}, 1(8):9.

\bibitem[{Raffel et~al.(2020)Raffel, Shazeer, Roberts, Lee, Narang, Matena,
  Zhou, Li, and Liu}]{t5}
Colin Raffel, Noam Shazeer, Adam Roberts, Katherine Lee, Sharan Narang, Michael
  Matena, Yanqi Zhou, Wei Li, and Peter~J. Liu. 2020.
\newblock \href {http://jmlr.org/papers/v21/20-074.html} {{Exploring the Limits
  of Transfer Learning with a Unified Text-to-Text Transformer}}.
\newblock \emph{Journal of Machine Learning Research}, 21(140):1--67.

\bibitem[{Ranasinghe et~al.(2021)Ranasinghe, Orasan, and
  Mitkov}]{qe_machine_translation}
Tharindu Ranasinghe, Constantin Orasan, and Ruslan Mitkov. 2021.
\newblock \href {https://doi.org/10.18653/v1/2021.acl-short.55} {{An
  Exploratory Analysis of Multilingual Word-Level Quality Estimation with
  Cross-Lingual Transformers}}.
\newblock In \emph{Proceedings of the 59th Annual Meeting of the Association
  for Computational Linguistics and the 11th International Joint Conference on
  Natural Language Processing (Volume 2: Short Papers)}, pages 434--440,
  Online. Association for Computational Linguistics.

\bibitem[{Rei et~al.(2021)Rei, Farinha, Zerva, van Stigt, Stewart, Ramos,
  Glushkova, Martins, and Lavie}]{rei-etal-2021-references}
Ricardo Rei, Ana~C Farinha, Chrysoula Zerva, Daan van Stigt, Craig Stewart,
  Pedro Ramos, Taisiya Glushkova, Andr{\'e} F.~T. Martins, and Alon Lavie.
  2021.
\newblock \href {https://aclanthology.org/2021.wmt-1.111} {{Are References
  Really Needed? Unbabel-{IST} 2021 Submission for the Metrics Shared Task}}.
\newblock In \emph{Proceedings of the Sixth Conference on Machine Translation},
  pages 1030--1040, Online. Association for Computational Linguistics.

\bibitem[{Rei et~al.(2020)Rei, Stewart, Farinha, and Lavie}]{comet}
Ricardo Rei, Craig Stewart, Ana~C Farinha, and Alon Lavie. 2020.
\newblock \href {https://doi.org/10.18653/v1/2020.emnlp-main.213} {{{COMET}: A
  Neural Framework for {MT} Evaluation}}.
\newblock In \emph{Proceedings of the 2020 Conference on Empirical Methods in
  Natural Language Processing (EMNLP)}, pages 2685--2702, Online. Association
  for Computational Linguistics.

\bibitem[{Reimers and Gurevych(2019)}]{sbert}
Nils Reimers and Iryna Gurevych. 2019.
\newblock \href {https://doi.org/10.18653/v1/D19-1410} {{Sentence-{BERT}:
  Sentence Embeddings using {S}iamese {BERT}-Networks}}.
\newblock In \emph{Proceedings of the 2019 Conference on Empirical Methods in
  Natural Language Processing and the 9th International Joint Conference on
  Natural Language Processing (EMNLP-IJCNLP)}, pages 3982--3992, Hong Kong,
  China. Association for Computational Linguistics.

\bibitem[{Reimers and Gurevych(2020)}]{distilscore}
Nils Reimers and Iryna Gurevych. 2020.
\newblock \href {https://doi.org/10.18653/v1/2020.emnlp-main.365} {{Making
  Monolingual Sentence Embeddings Multilingual using Knowledge Distillation}}.
\newblock In \emph{Proceedings of the 2020 Conference on Empirical Methods in
  Natural Language Processing (EMNLP)}, pages 4512--4525, Online. Association
  for Computational Linguistics.

\bibitem[{Scialom et~al.(2021{\natexlab{a}})Scialom, Dray, Lamprier,
  Piwowarski, Staiano, Wang, and Gallinari}]{scialom-etal-2021-questeval}
Thomas Scialom, Paul-Alexis Dray, Sylvain Lamprier, Benjamin Piwowarski, Jacopo
  Staiano, Alex Wang, and Patrick Gallinari. 2021{\natexlab{a}}.
\newblock \href {https://doi.org/10.18653/v1/2021.emnlp-main.529}
  {{{Q}uest{E}val: Summarization Asks for Fact-based Evaluation}}.
\newblock In \emph{Proceedings of the 2021 Conference on Empirical Methods in
  Natural Language Processing}, pages 6594--6604, Online and Punta Cana,
  Dominican Republic. Association for Computational Linguistics.

\bibitem[{Scialom et~al.(2019)Scialom, Lamprier, Piwowarski, and
  Staiano}]{scialom-etal-2019-answers}
Thomas Scialom, Sylvain Lamprier, Benjamin Piwowarski, and Jacopo Staiano.
  2019.
\newblock \href {https://doi.org/10.18653/v1/D19-1320} {{Answers Unite!
  Unsupervised Metrics for Reinforced Summarization Models}}.
\newblock In \emph{Proceedings of the 2019 Conference on Empirical Methods in
  Natural Language Processing and the 9th International Joint Conference on
  Natural Language Processing (EMNLP-IJCNLP)}, pages 3246--3256, Hong Kong,
  China. Association for Computational Linguistics.

\bibitem[{Scialom et~al.(2021{\natexlab{b}})Scialom, Martin, Staiano,
  de~La~Clergerie, and Sagot}]{scialom2021rethinking}
Thomas Scialom, Louis Martin, Jacopo Staiano, Eric~Villemonte de~La~Clergerie,
  and Beno{\^\i}t Sagot. 2021{\natexlab{b}}.
\newblock {Rethinking automatic evaluation in sentence simplification}.
\newblock \emph{arXiv preprint arXiv:2104.07560}.

\bibitem[{Sellam et~al.(2020)Sellam, Das, and Parikh}]{bleurt}
Thibault Sellam, Dipanjan Das, and Ankur Parikh. 2020.
\newblock \href {https://doi.org/10.18653/v1/2020.acl-main.704} {{{BLEURT}:
  Learning Robust Metrics for Text Generation}}.
\newblock In \emph{Proceedings of the 58th Annual Meeting of the Association
  for Computational Linguistics}, pages 7881--7892, Online. Association for
  Computational Linguistics.

\bibitem[{Specia et~al.(2010)Specia, Raj, and Turchi}]{specia2010machine}
Lucia Specia, Dhwaj Raj, and Marco Turchi. 2010.
\newblock {Machine translation evaluation versus quality estimation}.
\newblock \emph{Machine translation}, 24:39--50.

\bibitem[{Srikanth and Li(2021)}]{elaborative_simplification}
Neha Srikanth and Junyi~Jessy Li. 2021.
\newblock \href {https://doi.org/10.18653/v1/2021.findings-acl.455}
  {{Elaborative Simplification: Content Addition and Explanation Generation in
  Text Simplification}}.
\newblock In \emph{Findings of the Association for Computational Linguistics:
  ACL-IJCNLP 2021}, pages 5123--5137, Online. Association for Computational
  Linguistics.

\bibitem[{{\v{S}}tajner et~al.(2014){\v{S}}tajner, Mitkov, and
  Saggion}]{stajner-etal-2014-one}
Sanja {\v{S}}tajner, Ruslan Mitkov, and Horacio Saggion. 2014.
\newblock \href {https://doi.org/10.3115/v1/W14-1201} {{One Step Closer to
  Automatic Evaluation of Text Simplification Systems}}.
\newblock In \emph{Proceedings of the 3rd Workshop on Predicting and Improving
  Text Readability for Target Reader Populations ({PITR})}, pages 1--10,
  Gothenburg, Sweden. Association for Computational Linguistics.

\bibitem[{{\v{S}}tajner et~al.(2016){\v{S}}tajner, Popovic, Saggion, Specia,
  and Fishel}]{vstajner2016shared}
Sanja {\v{S}}tajner, Maja Popovic, Horacio Saggion, Lucia Specia, and Mark
  Fishel. 2016.
\newblock {Shared task on quality assessment for text simplification}.
\newblock \emph{Training}, 218(95):192.

\bibitem[{Sulem et~al.(2018{\natexlab{a}})Sulem, Abend, and
  Rappoport}]{sulem-etal-2018-bleu}
Elior Sulem, Omri Abend, and Ari Rappoport. 2018{\natexlab{a}}.
\newblock \href {https://doi.org/10.18653/v1/D18-1081} {{{BLEU} is Not Suitable
  for the Evaluation of Text Simplification}}.
\newblock In \emph{Proceedings of the 2018 Conference on Empirical Methods in
  Natural Language Processing}, pages 738--744, Brussels, Belgium. Association
  for Computational Linguistics.

\bibitem[{Sulem et~al.(2018{\natexlab{b}})Sulem, Abend, and
  Rappoport}]{sulem-etal-2018-semantic}
Elior Sulem, Omri Abend, and Ari Rappoport. 2018{\natexlab{b}}.
\newblock \href {https://doi.org/10.18653/v1/N18-1063} {{Semantic Structural
  Evaluation for Text Simplification}}.
\newblock In \emph{Proceedings of the 2018 Conference of the North {A}merican
  Chapter of the Association for Computational Linguistics: Human Language
  Technologies, Volume 1 (Long Papers)}, pages 685--696, New Orleans,
  Louisiana. Association for Computational Linguistics.

\bibitem[{Sun et~al.(2021)Sun, Jin, and Wan}]{doc_level_simplification}
Renliang Sun, Hanqi Jin, and Xiaojun Wan. 2021.
\newblock \href {https://doi.org/10.18653/v1/2021.emnlp-main.630}
  {{Document-Level Text Simplification: Dataset, Criteria and Baseline}}.
\newblock In \emph{Proceedings of the 2021 Conference on Empirical Methods in
  Natural Language Processing}, pages 7997--8013, Online and Punta Cana,
  Dominican Republic. Association for Computational Linguistics.

\bibitem[{Temnikova and Maneva(2013)}]{c-score}
Irina Temnikova and Galina Maneva. 2013.
\newblock \href {https://aclanthology.org/W13-2903} {The {C}-score {--}
  proposing a reading comprehension metrics as a common evaluation measure for
  text simplification}.
\newblock In \emph{Proceedings of the Second Workshop on Predicting and
  Improving Text Readability for Target Reader Populations}, pages 20--29,
  Sofia, Bulgaria. Association for Computational Linguistics.

\bibitem[{Thompson and Post(2020)}]{thompson-post-2020-automatic}
Brian Thompson and Matt Post. 2020.
\newblock \href {https://doi.org/10.18653/v1/2020.emnlp-main.8} {{Automatic
  Machine Translation Evaluation in Many Languages via Zero-Shot
  Paraphrasing}}.
\newblock In \emph{Proceedings of the 2020 Conference on Empirical Methods in
  Natural Language Processing (EMNLP)}, pages 90--121, Online. Association for
  Computational Linguistics.

\bibitem[{Vasilyev et~al.(2020)Vasilyev, Dharnidharka, and
  Bohannon}]{vasilyev-etal-2020-fill}
Oleg Vasilyev, Vedant Dharnidharka, and John Bohannon. 2020.
\newblock \href {https://doi.org/10.18653/v1/2020.eval4nlp-1.2} {{Fill in the
  {BLANC}: Human-free quality estimation of document summaries}}.
\newblock In \emph{Proceedings of the First Workshop on Evaluation and
  Comparison of NLP Systems}, pages 11--20, Online. Association for
  Computational Linguistics.

\bibitem[{Wan et~al.(2022)Wan, Liu, Yang, Zhang, Chen, Wong, and Chao}]{unite}
Yu~Wan, Dayiheng Liu, Baosong Yang, Haibo Zhang, Boxing Chen, Derek Wong, and
  Lidia Chao. 2022.
\newblock \href {https://doi.org/10.18653/v1/2022.acl-long.558} {{{U}ni{TE}:
  Unified Translation Evaluation}}.
\newblock In \emph{Proceedings of the 60th Annual Meeting of the Association
  for Computational Linguistics (Volume 1: Long Papers)}, pages 8117--8127,
  Dublin, Ireland. Association for Computational Linguistics.

\bibitem[{Wei et~al.(2022)Wei, Wang, Schuurmans, Bosma, brian ichter, Xia, Chi,
  Le, and Zhou}]{cot}
Jason Wei, Xuezhi Wang, Dale Schuurmans, Maarten Bosma, brian ichter, Fei Xia,
  Ed~H. Chi, Quoc~V Le, and Denny Zhou. 2022.
\newblock \href {https://openreview.net/forum?id=_VjQlMeSB_J} {Chain of thought
  prompting elicits reasoning in large language models}.
\newblock In \emph{Advances in Neural Information Processing Systems}.

\bibitem[{Wubben et~al.(2012)Wubben, van~den Bosch, and Krahmer}]{pbmt-r}
Sander Wubben, Antal van~den Bosch, and Emiel Krahmer. 2012.
\newblock \href {https://aclanthology.org/P12-1107} {{Sentence Simplification
  by Monolingual Machine Translation}}.
\newblock In \emph{Proceedings of the 50th Annual Meeting of the Association
  for Computational Linguistics (Volume 1: Long Papers)}, pages 1015--1024,
  Jeju Island, Korea. Association for Computational Linguistics.

\bibitem[{Xu et~al.(2016)Xu, Napoles, Pavlick, Chen, and Callison-Burch}]{sari}
Wei Xu, Courtney Napoles, Ellie Pavlick, Quanze Chen, and Chris Callison-Burch.
  2016.
\newblock \href {https://doi.org/10.1162/tacl_a_00107} {{Optimizing Statistical
  Machine Translation for Text Simplification}}.
\newblock \emph{Transactions of the Association for Computational Linguistics},
  4:401--415.

\bibitem[{Zhang et~al.(2020)Zhang, Kishore, Wu, Weinberger, and
  Artzi}]{bertscore}
Tianyi Zhang, Varsha Kishore, Felix Wu, Kilian~Q. Weinberger, and Yoav Artzi.
  2020.
\newblock \href {https://openreview.net/forum?id=SkeHuCVFDr} {{BERTScore:
  Evaluating Text Generation with BERT}}.
\newblock In \emph{International Conference on Learning Representations}.

\bibitem[{Zhang and Lapata(2017)}]{dress}
Xingxing Zhang and Mirella Lapata. 2017.
\newblock \href {http://aclweb.org/anthology/D17-1063} {{Sentence
  Simplification with Deep Reinforcement Learning}}.
\newblock In \emph{Proceedings of the 2017 Conference on Empirical Methods in
  Natural Language Processing}, pages 595--605. Association for Computational
  Linguistics.

\bibitem[{Zhang et~al.(2023)Zhang, Cui, Zhao, Bi, and Shi}]{zhang-et-al}
Yue Zhang, Leyang Cui, Enbo Zhao, Wei Bi, and Shuming Shi. 2023.
\newblock \href {http://arxiv.org/abs/2310.07299} {{RobustGEC: Robust
  Grammatical Error Correction Against Subtle Context Perturbation}}.

\bibitem[{Zhao et~al.(2020)Zhao, Glava{\v{s}}, Peyrard, Gao, West, and
  Eger}]{xmoverscore}
Wei Zhao, Goran Glava{\v{s}}, Maxime Peyrard, Yang Gao, Robert West, and
  Steffen Eger. 2020.
\newblock \href {https://www.aclweb.org/anthology/2020.acl-main.151} {{On the
  Limitations of Cross-lingual Encoders as Exposed by Reference-Free Machine
  Translation Evaluation}}.
\newblock In \emph{Proceedings of the 58th Annual Meeting of the Association
  for Computational Linguistics}, pages 1656--1671, Online. Association for
  Computational Linguistics.

\bibitem[{Zhao et~al.(2023)Zhao, Durmus, and Yeung}]{bets}
Xinran Zhao, Esin Durmus, and Dit-Yan Yeung. 2023.
\newblock \href {https://doi.org/10.18653/v1/2023.findings-acl.838} {Towards
  reference-free text simplification evaluation with a {BERT} {S}iamese network
  architecture}.
\newblock In \emph{Findings of the Association for Computational Linguistics:
  ACL 2023}, pages 13250--13264, Toronto, Canada. Association for Computational
  Linguistics.

\end{thebibliography}

\section{Language Resource References}
\label{lr:ref}
\bibliographystylelanguageresource{lrec-coling2024-natbib}
\bibliographylanguageresource{language_resources_references}

\appendix
\section{Implementation Details}\label{sec:implementation_details}

\subsection{Data}
For the first pretraining stage, we use source sentences from the first 10 volumes of the OpenWebText dataset. We use the {\tt muss\_en\_wikilarge\_mined} checkpoint for the MUSS model~\cite{muss}. For the GPT models, we randomly sample five source-simplification pairs from the TurkCorpus dataset \cite{turkcorpus} as in-context examples. We use the following prompt template: 
\begin{itemize}
    \item System prompt: {\em You are a helpful assistant that simplifies English sentences, making them easier to read while preserving key meanings.}
    \item User prompt: {\em Follow the examples and simplify the sentence, making it easier to read while preserving key meanings. Reply with only the simplified sentence. Sentence:} \{...\} {\em Simplification:} \{...\} {\em ... Sentence:} \{...\} 
\end{itemize}
with the system prompt only applicable to the {\tt GPT-3.5-turbo} model. For the models in the second pretraining stage, we use the model outputs as published by their authors. 

For the pretraining stages, we randomly select 40\% of the system outputs for augmentation. Each instance selected for augmentation is assigned to be augmented by deletion, scrambling or by swapping the complex and simplified sentences respectively with probabilities of 0.3, 0.3 and 0.4. We use the {\tt NLTK Tree Bank Word Tokenizer}\footnote{\url{https://www.nltk.org/api/nltk.tokenize.TreebankWordTokenizer.html}} and uniformly randomly select one to four words for deletion and one to five words for scrambling. 

We use the HuggingFace {\tt Evaluate} implementation\footnote{\url{https://huggingface.co/spaces/evaluate-metric/bleu}} of BLEU, the Sentence Transformer implementation of {\tt all-distillroberta-v1}\footnote{\url{sentence-transformers/all-distilroberta-v1}} for SBERT, the HuggingFace {\tt Evaluate} implementation\footnote{\url{https://huggingface.co/spaces/evaluate-metric/perplexity}} of GPT-2 perplexity and the EASSE~\cite{easse} implementation of SARI. For FKGL, we use the syllable and lexicon counts calculated using the {\tt Textstat} package.\footnote{\url{https://pypi.org/project/textstat/}} Each supervision signal is normalized across the dataset.

\subsection{Model Implementation and Training}

We base \textsc{REFeREE} on the HuggingFace implementation \footnote{\url{https://huggingface.co/microsoft/deberta-v3-base}} of the {\tt DeBERTa-v3-base} model with 12 layers and a hidden size of 768. The model takes as input the source and simplified sentences delimited by a {\tt <SEP>} token. We use the embedding corresponding to the {\tt <BOS>} token as the sequence embedding and feed it into separate linear regression heads for each supervision signal. 

For each stage, the model is trained on the unweighted sum of L2 losses from the training signals. We use the Adam optimizer~\cite{adam} with $\epsilon = 10^{-6}$, $\beta_1 = 0.9$, and $\beta_2 = 0.999$. For the three stages, we respectively train the model for a maximum of three, 30, and 50 epochs with a learning rate of $10^{-5}$, $10^{-5}$, and $10^{-7}$ and apply early-stopping based on the loss on development sets.

For the fine-tuned BLEURT model, we start from the {\tt BLEURT-20-D12} checkpoint\footnote{\url{https://github.com/lucadiliello/bleurt-pytorch}} fine-tuned on WMT and reinitialize the regression head. The fine-tuning follows the same process as \textsc{REFeREE}.

\end{document}